\documentclass[10pt,twocolumn,letterpaper]{article}

\usepackage{cvpr}
\usepackage{times}
\usepackage{epsfig}
\usepackage{graphicx}
\usepackage{amsmath}
\usepackage{amssymb}

\usepackage[utf8]{inputenc} % allow utf-8 input

\usepackage{url}            % simple URL typesetting
\usepackage{booktabs}       % professional-quality tables
\usepackage{amsfonts}       % blackboard math symbols
\usepackage{nicefrac}       % compact symbols for 1/2, etc.
\usepackage{microtype}      % microtypography
\usepackage{caption}
\usepackage{colortbl}
\usepackage{graphicx}
\usepackage{subcaption}
\usepackage{float}
\usepackage{gensymb}
\usepackage{subfiles}
\usepackage{indentfirst}
\usepackage{multirow}
\usepackage{bm}
\usepackage[symbol]{footmisc}

\usepackage[breaklinks=true,bookmarks=false]{hyperref}

\cvprfinalcopy % *** Uncomment this line for the final submission

\graphicspath{{images/}{../images/}}

\newcommand{\aadd}{\textsc{A2D2}}
\newcommand{\aaddlong}{Audi Autonomous Driving Dataset}
\newcommand{\aaddurl}{\url{https://a2d2-dataset.github.io}}
\newcommand{\aaddlink}{\href{https://a2d2-dataset.github.io}{\textbf{A2D2}}}
\newcommand{\cities}{Gaimersheim, Munich, and Ingolstadt}
\newcommand{\region}{the south of Germany}
\newcommand{\car}{Audi Q7 e-tron}

\hypersetup{
  dvipdfm,    %put here the correct(!) driver you are using
  colorlinks=true,    %no frame around URL
  urlcolor=blue,    %no colors
  menucolor=black,    %no colors
  linkcolor=black,    %no colors
  pagecolor=black,    %no colors
  bookmarks=true,    %tree-like TOC
  bookmarksopen=true,    %expanded when starting
  hyperfootnotes=false,    %no referencing of footnotes, does not compile
  pdfpagemode=UseOutlines    %show the bookmarks when starting the pdf viewer
}

\begin{document}
\title{A2D2: Audi Autonomous Driving Dataset}

\author{%
Jakob Geyer\qquad
Yohannes Kassahun\qquad
Mentar Mahmudi\qquad
Xavier Ricou\qquad
Rupesh Durgesh\qquad \\
Andrew S. Chung\qquad 
Lorenz Hauswald\qquad
Viet Hoang Pham \qquad
Maximilian M\"uhlegg\qquad 
Sebastian Dorn \qquad \\
Tiffany Fernandez\textsuperscript{*} \qquad
Martin J\"anicke \qquad
Sudesh Mirashi\textsuperscript{*} \qquad
Chiragkumar Savani \qquad
Martin Sturm \qquad \\
Oleksandr Vorobiov\textsuperscript{*} \qquad
Martin Oelker \qquad
Sebastian Garreis \qquad
Peter Schuberth\qquad \\
\\
Audi AG \\
{\tt\small a2d2-dataset@googlegroups.com}
}

%\institute{Audi AG}

\twocolumn[{%
  \renewcommand\twocolumn[1][]{#1}%
  \maketitle
  \begin{center}
    \vspace{-6mm}
    \centering
    \includegraphics[width=1.0\textwidth]{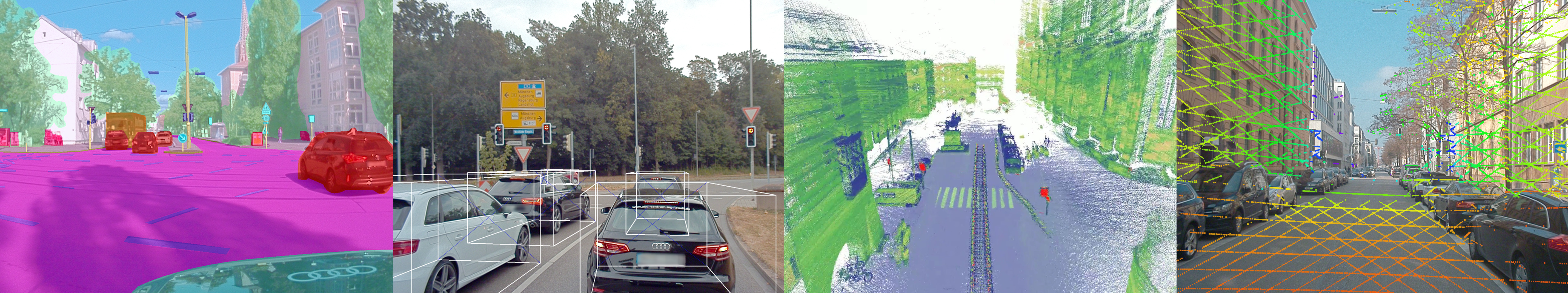}
    \captionof{figure}{Visualizations of \aadd{} data.
      From left: semantic segmentation, 3D bounding boxes,
      dense point cloud from SLAM, single frame point cloud
	  overlaid on corresponding camera image.
	  %\textbf{---mentar: stronger overlay needed for the 4th Fig.}
  		}
    \label{fig:headline}
  \end{center}%
}]

\footnotetext[1]{Work done while at Audi AG}
  \begin{abstract}

    Research in machine learning, mobile robotics, and autonomous driving is accelerated by the availability of high quality annotated data.
    To this end, we release the \aaddlong{} (\aadd{}).
    Our dataset consists of simultaneously recorded images and 3D point clouds, together with 3D bounding boxes, semantic segmentation, instance segmentation, and data extracted from the automotive bus.
    Our sensor suite consists of six cameras and five LiDAR units, providing full 360{\degree} coverage.
    The recorded data is time synchronized and mutually registered.
    Annotations are for non-sequential frames: 41,277 frames with semantic segmentation image and point cloud labels, of which 12,497 frames also have 3D bounding box annotations for objects within the field of view of the front camera.
    In addition, we provide 392,556 sequential frames of unannotated sensor data for recordings in three cities in \region{}.
    These sequences contain several loops.
    Faces and vehicle number plates are blurred due to GDPR legislation and to preserve anonymity.
    \aadd{} is made available under the CC BY-ND 4.0 license, permitting commercial use subject to the terms of the license.
    Data and further information are available at \aaddurl{}.
    %\keywords{3D Point Clouds -- Datasets and Evaluation -- Deep Learning: Applications, Methodology, and Theory -- Recognition: Detection, Categorization, Retrieval and Matching -- Segmentation, Grouping and Shape -- Vision for Robotics}
  \end{abstract}

  \section{Introduction}
  \label{sec:intro}

  Access to high quality data has proven crucial to the development of autonomous driving systems.
  %Many current datasets for autonomous driving are primarily focused on camera-based data, restricting the scope of systems to which they are relevant.
  In this paper we present the \aaddlong{} (\aadd{}) which provides camera, LiDAR, and vehicle bus data, allowing developers and researchers to explore multimodal sensor fusion approaches.

  While some datasets such as KITTI \cite{geiger2013} and ApolloScape \cite{wang2019apolloscape} also provide both LiDAR and camera data, more recent datasets \cite{nuscenes,waymo2} have put an emphasis on providing full surround sensor coverage.
  We believe this is important for further advances in autonomous driving, and therefore have released \aadd{} with full surround camera and LiDAR data.
  We also include vehicle bus data, which provides additional information about car state (\eg translational/rotational speed and acceleration, steering wheel angle, throttle, brake, etc.).

  The vast majority of publicly available datasets are released under licenses permitting research-only use.
  Whilst we understand the reasons for this, we want to push progress in the field by publishing \aadd{} under the less restrictive CC BY-ND 4.0 license, which allows commercial use (subject to the terms of the license).
  We hope this will help researchers, particularly those working within commercial enterprises.
  %Furthermore, by adhering to the recent GDPR regulations, we hope \aadd{} will be readily utilizable by all parties.

  By releasing \aadd{}, we seek to
  \begin{enumerate}
  \item[a)] catalyse research in machine learning and robotics, especially research related to autonomous driving
  \item[b)] provide a public dataset from a realistic autonomous driving sensor suite
  \item[c)] engage with the wider research community
  \item[d)] contribute to startups and other commercial entities by freely releasing data which is expensive to generate.
  \end{enumerate}

  In summary, we release \aadd{} to foster research, in keeping with our ethos of promoting innovation and actively participating in the research community.

  Our main contributions are as follows:

  \begin{itemize}
    \item The setup and calibration of our \car{} data collection platform is discussed in Section \ref{sec:Dataset}.
    We consider this platform to be broadly comparable to many autonomous driving development platforms currently in use by commercial entities.

    \item We provide the community with a commercial grade driving dataset, suitable for many perception tasks. It includes extensive vehicle bus data, which has hitherto been lacking in public datasets.

    \item We evaluate the performance of a semantic segmentation convolutional network on \aadd{} in Section \ref{sec:experiments}.

    \item We release \aadd{} under the commercial-friendly CC BY-ND 4.0 license.

  \end{itemize}

  \section{Related Work}
  \label{sec:rworks}

  This section provides a brief survey of datasets relevant to the development of autonomous driving systems.
  We focus on the most comparable recent datasets, which strongly emphasize multimodal sensor data.
  We present them in chronological order.
  
  \subsection{Datasets}

  The KITTI \cite{geiger2013} dataset was a pioneer in the field which, together with its associated benchmarks \cite{Geiger2012}, has been highly influential.
  The data collection vehicle was equipped with four video cameras (two color, two grayscale), a 3D laser scanner, and a GPS/IMU inertial navigation system.
  Several challenges were published on tasks such as 2D and 3D object detection, SLAM, depth prediction, tracking, and optical flow.

  As the viability of image semantic segmentation solutions for autonomous driving increased, so too did the need for relevant semantically labeled images.
  The Cityscapes dataset \cite{cordts2016} sought to address this.
  The data were collected in 50 cities in Germany, and recorded in dense urban traffic.
  The scenes were captured using stereo-pair color cameras and were annotated semantically on both instance and pixel level. 
  Cityscapes contains 5000 fine labeled images spanning over 30 classes.

  The Mapillary Vistas dataset \cite{neuhold2017} also provides semantic segmentation labels for urban, rural, and off-road scenes.
  The dataset contains 25,000 densely annotated street-level images from locations around the world.
  The dataset is heterogeneous in that the capture devices span mobile phones, tablets, and assorted cameras.

  ApolloScape \cite{wang2019apolloscape} is a large dataset consisting of over 140,000 video frames from various locations in China under varying weather conditions.
  Pixel-wise semantic annotation of the recorded data is provided in 2D, with point-wise semantic annotation in 3D for 28 classes.
  In addition, the dataset contains lane marking annotations in 2D.
  To our knowledge, ApolloScape is the largest publicly available semantic segmentation dataset for autonomous driving applications.

  The Berkeley Deep Drive dataset (BDD-100k) \cite{yu2018} has a stronger emphasis on 2D bounding boxes but also contains pixel-wise segmentation annotations for 10,000 images.
  BDD-100K contains 100,000 images with 2D bounding boxes as well as street lanes, markings, and traffic light color identification.

  Several semantic point cloud datasets have also been made available, \eg \cite{Munoz2009ContextualCW,steder2011,behley2012,zhang2015,hackel2017}.  
  However, SemanticKITTI \cite{behley2019iccv} is significantly larger than its predecessors.
  It also has the advantage of being based on the already widely-used KITTI dataset, providing point-wise semantic annotations of all 22 pointclouds of the KITTI Vision Odometry Benchmark.
  This corresponds to 43,000 separately annotated scans.

  KITTI highlighted the importance of multi-modal sensor setups for autonomous driving, and the latest datasets have put a strong emphasis on this aspect.
  nuScenes \cite{nuscenes} is a recently released dataset which is particularly notable for its sensor multimodality.
  It consists of camera images, LiDAR point clouds, and radar data, together with 3D bounding box annotations.
  It was collected during the day and night under clear weather conditions. 
  
  The Lyft Level 5 AV Dataset~\cite{lyft2019} has camera and LiDAR data in the nuScene data format.
  It has a strong emphasis on 3D bounding box detection and tracking.

  Most recently, the Waymo Open Dataset~\cite{waymo2} was released with 12 million 3D bounding box annotations on LiDAR point clouds and 1.2 million 2D bounding box annotations on camera frames.
  In total it consists of 1000 20-second sequences in urban and suburban scenarios, under various weather and lighting conditions.

  \subsection{Comparison}
  \label{sec:Comparison}
    \begin{footnotesize}
  	\begin{table*}[h]
  		\centering
  		\begin{footnotesize}
  			\begin{tabular}{m{1.3cm} m{2cm} m{2cm} m{2cm} m{2cm} m{2cm} m{2cm} m{2cm} m{2cm} m{2cm}}
  			\setlength\tabcolsep{6 pt}
  			%\begin{tabular}{m{1.2cm} m{1.5cm} m{1.5cm} m{1.5cm} m{1.5cm} m{1.5cm} m{1.5cm} m{1.5cm} m{1.5cm} m{1.5cm}}
  				&\href{http://www.cvlibs.net/datasets/kitti/}{\textbf{KITTI}}
  				&\href{http://apolloscape.auto/index.html}{\textbf{Apollo Scape}}
  				&\href{https://www.nuscenes.org/}{\textbf{nuScenes}}
  				&\href{https://level5.lyft.com/dataset/}{\textbf{Lyft Level 5}}
  				&\href{https://waymo.com/open/}{\textbf{Waymo OD}}
  				&\aaddlink{} \\
  				\hline
  				\textbf{Cameras}      & 4 (0.7MP)          & 2 (9.2MP) & 6 (1.4MP) & 7 (2.1MP)\small\textsuperscript{$\star$}  & 3 (2.5MP) + 2 (1.7MP) & 6 (2.3MP)\\
  				\hline
  				\textbf{LiDAR sensors}    & 1 (64 channel)       &   2 (N/A)  &  1 (32 channel) & 1 + 2 aux. (40 channel)  & 1 + 4 aux. (64 channel\small\textsuperscript{$\dagger$}) & 5 (16 channel) \\
  				\hline
  				\textbf{Vehicle bus data}  & GPS+IMU\textsuperscript{$\ddagger$}  & GPS+IMU\textsuperscript{$\ddagger$} & - & - & velocity, angular velocity & GPS, IMU, steering angle, brake, throttle, odometry, velocity, pitch, roll \\
  				\hline
  				\textbf{Location}      & urban, one city   & urban, various cities& urban, two cities & urban & urban & urban, highways, country roads, three cities \\
  				\hline
  				\textbf{Hours}        & day       & day & day, night & day & day, night & day \\
  				\hline
  				\textbf{Weather}      & sunny, cloudy   & various weather & various weather & various weather & various weather & various weather \\
  				\hline
  				\textbf{Objects}      & 3D     & pixel, 3D semantic points & 3D & 3D & 3D, 2D & 3D, pixel \\
  				\hline
  				\textbf{Last updated}    & 2015 & 2018 & 2019 & 2019 & 2019 & 2020 \\  \hline
  				
  			\end{tabular}
  		\end{footnotesize}
  		
  		\caption{Comparison of datasets with multimodal sensor approach.~\textsuperscript{$\star$}The following alternative setup was also used: 6 ($1224 \times 1024$) + 1 ($2048 \times 864$).~\textsuperscript{$\dagger$}Number of channels refers to main LiDAR. \textsuperscript{$\ddagger$}Provided by additional sensors}\label{tbl:comparison}
  		
  	\end{table*}
  \end{footnotesize}

  We compare \aadd{} to the other multimodal datasets listed in Table \ref{tbl:comparison}.
  All contain LiDAR point clouds and images from several cameras. % , while nuScenes and Lyft Level 5 additionally include vehicle bus data.
  Most focus on object detection for autonomous shuttle fleets operating in predefined urban scenarios. 

  \aadd{} similarly contains images and point clouds, but, in addition, it provides extensive vehicle bus data including steering wheel angle, throttle, and braking.
  This allows \aadd{} to be used for more fields of research in autonomous driving, \eg end-to-end learning as in \cite{bojarski2016} and \cite{xiao2019multimodal,haavaldsen2019autonomous} (synthetic data).
  To the best of our knowledge other multimodal datasets do not provide such data.
  
  Since other datasets focus on object detection, their LiDAR setups are configured so that the highest detected points are slightly above the recording vehicle.
  In contrast, the scan patterns of the five LiDAR setup (80 channels in total) used in \aadd{} are optimized for uniform distribution and maximum overlap with the camera frames.
  As a result they also cover a large area above the vehicle and capture large static objects such as high buildings.
  This makes the dataset particularly relevant for SLAM and 3D map generation, \eg \cite{loopclosure2dslam,alismail2014,kohlbrecher2011}.
  
  \aadd{} complements current multimodal datasets by having a stronger emphasis on semantic segmentation and vehicle bus data. Furthermore, the unannotated sequences
  focus on longer consecutive LiDAR and camera data suitable for self-supervised approaches.

	\section{Dataset} \label{sec:Dataset}
	\aadd{} includes data recorded on highways, country roads, and cities in \region{}.
	The data were recorded under cloudy, rainy, and sunny weather conditions.
	We provide semantic segmentation labels, instance segmentation labels, and 3D bounding boxes for non-sequential frames: 41,277 images have semantic and instance segmentation labels for 38 categories.
	All images have corresponding LiDAR  point clouds, of which 12,497 are annotated with 3D bounding boxes within the field of view of the front-center camera.
	We also provide unannotated sequence data.

	\subsection{Data Collection Platform}
	\label{sec:Data_Collection_Platform}
\begin{figure*}
		\centering
		\begin{subfigure}[b]{0.26\textwidth}
			\includegraphics[width=\textwidth]{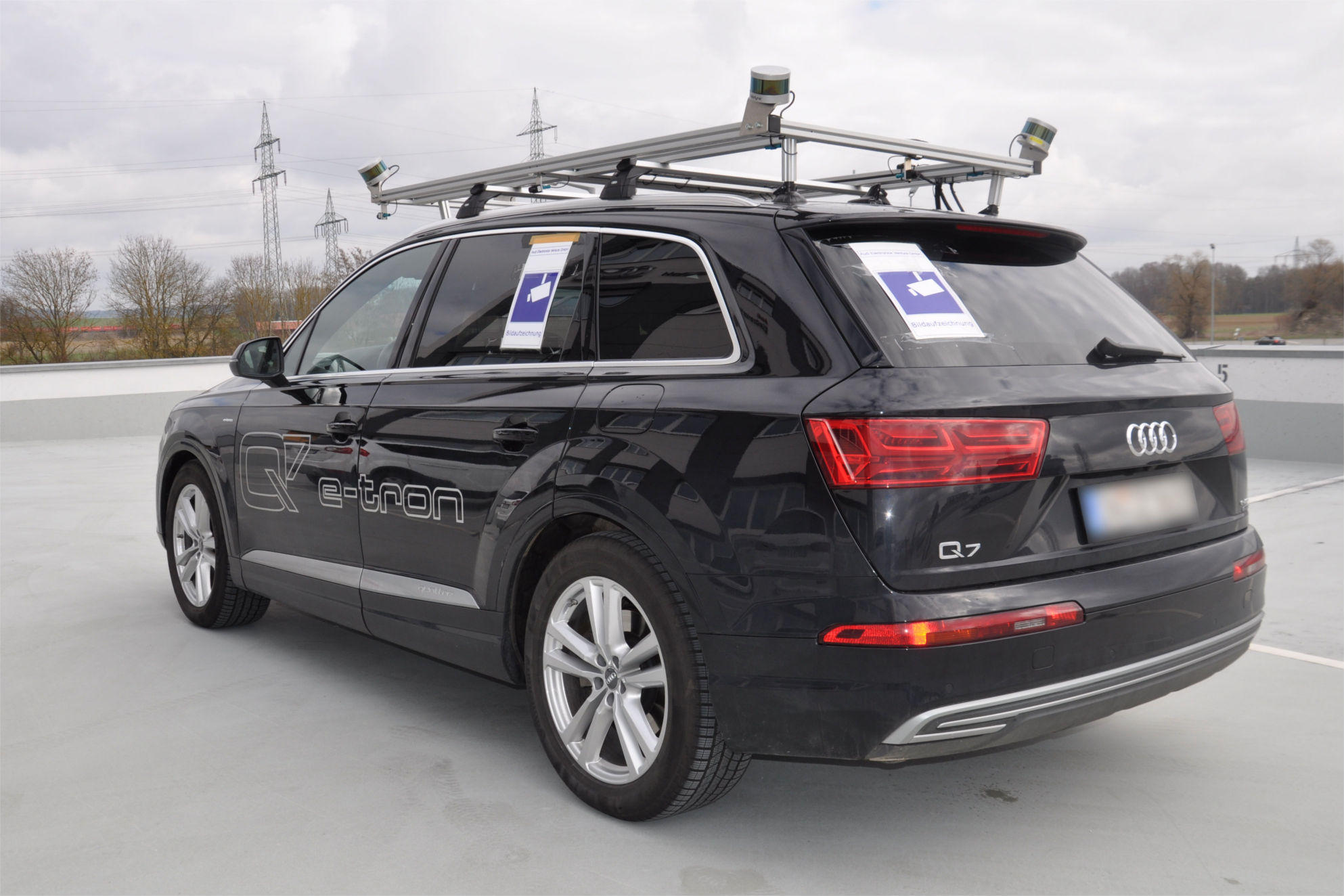}
			\vspace*{0ex}
			\caption{\car{}}
			\label{fig:Q7}
		\end{subfigure}
		\begin{subfigure}[b]{0.26\textwidth}
			\hspace*{2ex}
			\includegraphics[width=0.9\textwidth]{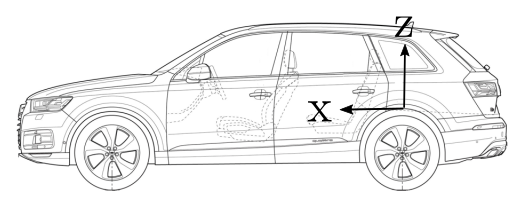}
			\vspace*{5ex}
			\caption{Global reference frame}
			\label{fig:GlobalRefFrame}
		\end{subfigure}
		\begin{subfigure}[b]{0.44\textwidth}
			\includegraphics[width=\linewidth]{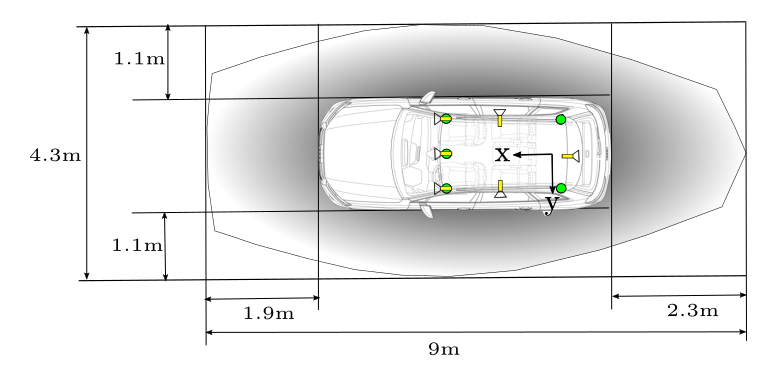}
			\caption{LiDAR blind spots (shaded).
				Sensor placement is shown in yellow (camera) and green (LiDAR)}
			\label{fig:blindspot}
		\end{subfigure}
		\caption{}
		\label{fig:audiq7}
	\end{figure*}
	We collected data using an \car{} equipped with six cameras and five Velodyne VLP-16 sensors (see Tables \ref{tbl:sensors_mounted}, \ref{tbl:sekonix_cameras}, and \ref{tbl:velodyne_lidars}).
	In addition to the camera and LiDAR data from our sensor suite, we also recorded the vehicle bus data.
	Figure \ref{fig:audiq7}(a) shows the vehicle used for data collection.
	% ac: camera frame rate, bit depth, etc.
	% ac: final sensor pose
	% ac: formats RAW->YUV 12-bit -> RGB (8bpp) via FFMPEG
	The sensor configuration and the frame of reference $\bm{g}$ are visualized in Figure \ref{fig:audiq7}(b).
%	The $x$-axis points forward, the $y$-axis points left, passing through the highest points on the rear wheel arches, and the $z$-axis points up.
	The $y$-axis passes through the highest points on the rear wheel arches.
	The poses of the camera and LiDAR sensors are given with respect to this frame of reference.

	\begin{footnotesize}
		\begin{table}[h]
			\centering
			\begin{footnotesize}
				\begin{tabular}{rll}
					%\hline
					Sensor	& Location		&  Type\\
					\hline %\hline
					Camera	& Front-center	& Sekonix SF3325-100\\
					Camera	& Front-left	& Sekonix SF3324-100\\
					Camera	& Front-right	& Sekonix SF3324-100\\
					Camera	& Side-left		& Sekonix SF3324-100\\
					Camera	& Side-right	& Sekonix SF3324-100\\
					Camera	& Rear-center	& Sekonix SF3324-100\\
					\midrule
					LiDAR	& Front-center	& Velodyne VLP-16\\
					LiDAR	& Front-left	& Velodyne VLP-16\\
					LiDAR	& Front-right	& Velodyne VLP-16\\
					LiDAR	& Rear-left		& Velodyne VLP-16\\
					LiDAR	& Rear-right	& Velodyne VLP-16\\
					\hline
				\end{tabular}
			\end{footnotesize}
			\caption{Sensor suite}
			\label{tbl:sensors_mounted}
		\end{table}
		\begin{table}[h]
			\centering
			\begin{footnotesize}
				\begin{tabular}{rll}
					%\hline
					& SF3225-100	&  SF3224-100\\
					\hline %\hline
					Horizontal FOV 	& 60{\degree}	& 120{\degree}\\
					Vertical FOV	& 38{\degree}	& 73{\degree}\\
					Diagonal FOV	& 70{\degree}	& 146{\degree}\\
					Sensor			& Onsemi AR0231	& Onsemi AR0231\\
					Resolution 		& 1928 x 1208 (2MP)& 1928 x 1208 (2MP)\\
					Colour Filter Array & RCCB 		& RCCB\\
					\hline
				\end{tabular}
			\end{footnotesize}
			\caption{Camera specifications}
			\label{tbl:sekonix_cameras}
		\end{table}
		\begin{table}[h]
			\centering
			\begin{footnotesize}
				\begin{tabular}{rll}
					%\hline
					& VLP-16\\
					\hline %\hline
					Azimuthal FOV 	& 360{\degree}\\
					Vertical FOV	& 30{\degree} (+15{\degree} to -15{\degree})\\
					Channels 		& 16\\
					Vertical resolution & 2{\degree}\\
					Frequency 		& 5-20Hz (10Hz used for \aadd{})\\
					Range			& up to 100m\\
					Rate			& up to $\sim$300,000 points/second\\
					\hline
				\end{tabular}
			\end{footnotesize}
			\caption{LiDAR specifications \cite{velodyne}} %VLP-16 datasheet
			\label{tbl:velodyne_lidars}
		\end{table}
	\end{footnotesize}

	\subsubsection{Sensor Setup}
	\label{sec:PoseOptimisation}
	We chose to mount the sensors on the roof of the vehicle with the aim of obtaining $360\degree$ environmental coverage, and to be symmetric with respect to the $x$-$z$-plane.
	The number of sensors was limited by data recording bandwidth.
	Three fisheye (120\degree~horizontal FOV) cameras provide views to the left, right, and rear of the vehicle.
	More emphasis was put on the front view which was covered with three cameras: a pair of fisheye cameras  mounted on the front-left and front-right of the roof, and a rectilinear (60\degree~horizontal FOV) camera mounted between them to provide a more detailed and less distorted view.
	LiDAR sensors were placed at each corner of the setup and above the front center camera.
	Figure \ref{fig:audiq7}(c) shows the described sensor placement.
	
	After fixing the sensor locations, the camera and LiDAR sensor orientations were optimized manually by visualizing the covered 3D region in CAD software.
	The goal of this process was to minimize the blind spot around the vehicle while maximizing camera and LiDAR field of view overlap.
	Figure \ref{fig:audiq7}(c) depicts the blind spot of the LiDAR sensors, evaluated at the ground plane.
	Outside the LiDAR blind spot, the fields of view of the cameras and the LiDAR sensors largely overlap (over 90 \%).
	This is demonstrated visually in the upper panel of Figure \ref{fig:MappingLiDARPoints}.

	\subsubsection{Sensor Calibration/Registration}
	\label{sec:CalibrationRegistration}
	The sensors were mounted on the vehicle as detailed in Section \ref{sec:PoseOptimisation} and measurements were made of their poses.
	In-situ calibration was still required, especially to account for error in measuring sensor angles.
	
	Firstly, the measured pose of the front-center LiDAR sensor with respect to the reference frame was assumed to be accurate. Using it as a reference, all remaining sensor poses were determined relative to this LiDAR.
	
	Secondly, we performed LiDAR to LiDAR mapping.
	To do so, LiDAR data was recorded in a static environment, with no dynamic objects.
	During data collection the vehicle was also stationary.
	An ICP-based registration \cite{beslmckay1992} was performed for determining the relative pose of the LiDAR sensors with respect to each other. %Using the measured pose of the front-center LiDAR, the poses with respect to the reference frame were obtained.
	
	Thirdly, intrinsic camera calibration was performed for all cameras using 2D patterns (checkerboards).
	
\begin{figure*}[ht]
		\centering
		\includegraphics[width=0.76\linewidth]{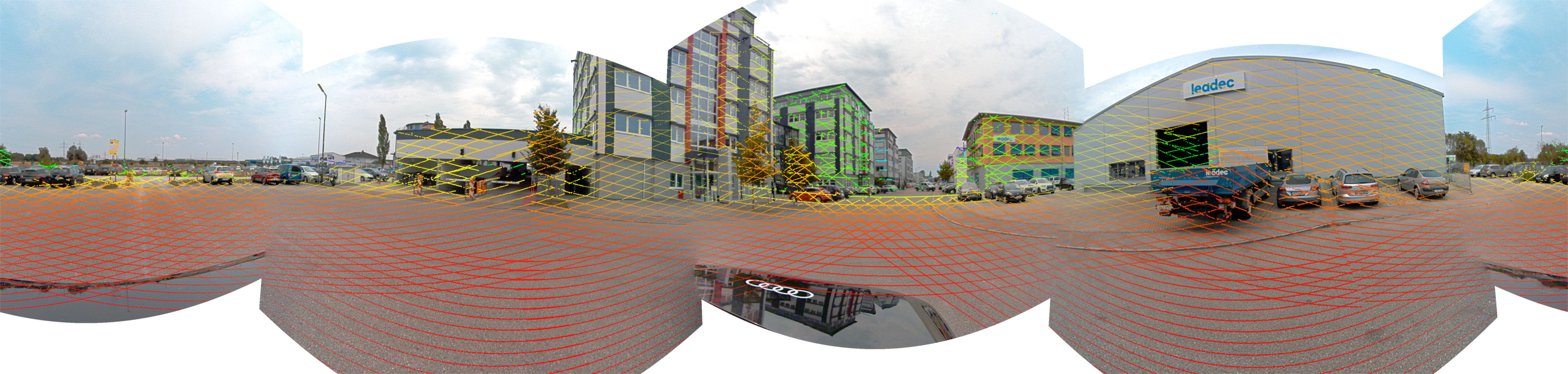}
		\begin{tabular} {cc}
			\includegraphics[width=0.36\linewidth]{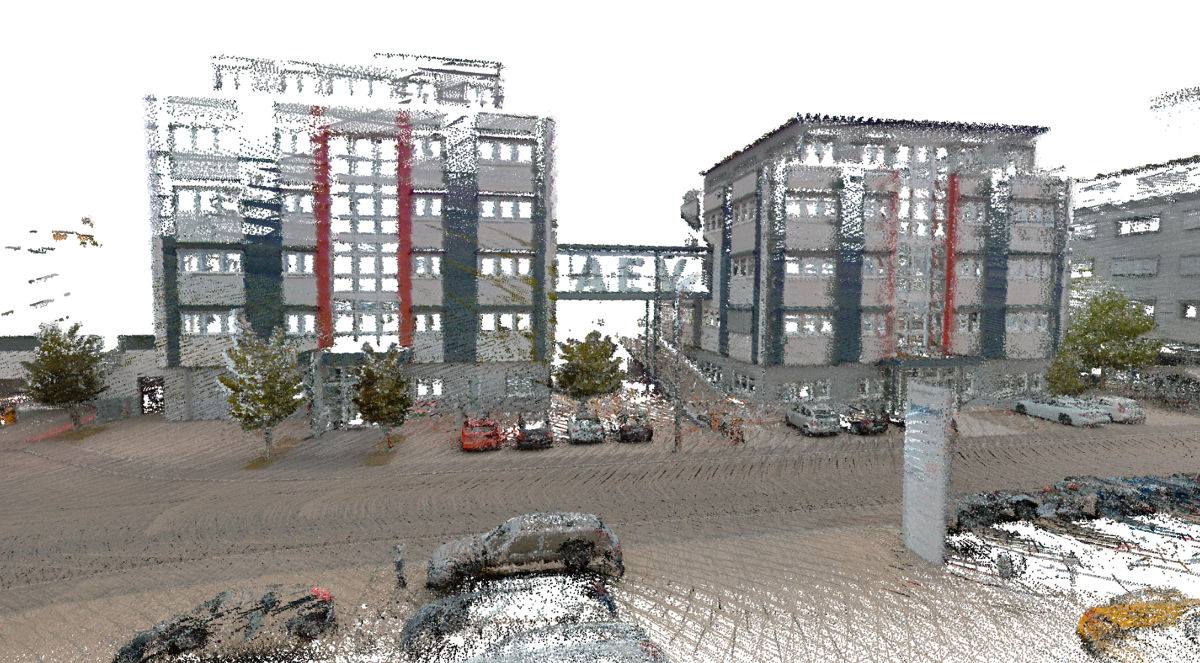}
		& \hspace*{-1.4ex}
			\includegraphics[width=0.36\linewidth]{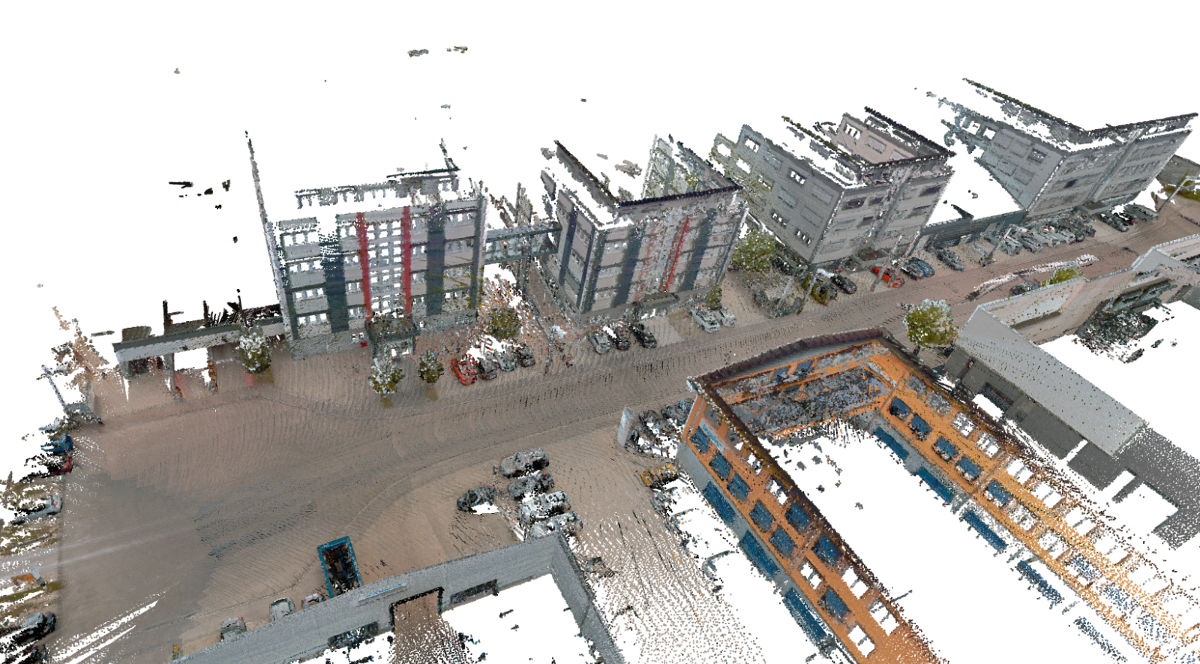}
		\end{tabular}
		\caption{Upper panel: Mapping of LiDAR points onto stitched camera images. Lower panels: colored 3D maps generated from a 30 second sequence using an ICP-based open-loop SLAM.}
		\label{fig:MappingLiDARPoints}
	\end{figure*}

	Finally, the LiDAR to camera mapping was determined as follows:
	Using a recording with our data recording vehicle in motion (see Section \ref{sec:SequenceData}), a large combined LiDAR point cloud was computed using ego-motion correction based on the bus data and then an ICP-based open-loop SLAM \cite{lu1997robot} (see lower panel of Figure \ref{fig:MappingLiDARPoints}).
	Given extrinsic camera parameters and using interpolation, a depth image as seen by the respective camera can be computed.
	Keeping the measured camera positions fixed, the camera angles were determined by optimizing for edge correspondence in the depth and RGB camera images.
	The top panel of Figure \ref {fig:MappingLiDARPoints} shows LiDAR points projected onto camera images using the resulting mappings.

	The result of the calibration is given in a configuration file containing:

\begin{itemize}
	\item a view, $\bm{v}$, for each sensor as well as the global frame of reference of the vehicle.
	Each view describes the pose of a sensor with respect to the reference frame, and is given by a tuple of three vectors so that $\bm{v} = (\bm{o}, \bm{x}, \bm{y})$.
	The vector $\bm{o} \in \mathbb{R}^{3}$ specifies the Cartesian coordinates (in meters) of the origin of the sensor.
	$\bm{x}, \bm{y} \in \mathbb{R}^{3}$ are unit vectors pointing in the direction (in Cartesian coordinates) of the sensor's $x$-axis and the $y$-axis.
	The $z$-axis of the sensor completes an orthonormal basis and thus is obtained by the cross-product $\bm{z} = \bm{x} \times \bm{y}$.
	\item the following parameters for each camera sensor:
	\begin {itemize}
	\item $\bm{K} \in \mathbb{R}^{3 \times 3}$ -- the intrinsic camera matrix of the original (distorted) camera image
	\item $\bm{M} \in \mathbb{R}^{3 \times 3}$ -- the intrinsic camera matrix of the undistorted camera image
	\item $\bm{d} \in \mathbb{R}^{4}$ -- distortion parameters of the original (distorted) camera image
	\item $\bm{r} \in \mathbb{N}^{2}$ -- resolution (columns, rows) of undistorted as well as original camera image
	\item $L \in \{\mbox{Fisheye}, \mbox{Wide-angle}\}$ -- type of lens.
	\end {itemize}
\end{itemize}
%In order to compare images from, e.g., two different camera sensors, it is convenient to rotate and shift the respective views $\bm{v}_1, \bm{v}_2$ into each other. The corresponding (homogeneous) transformation matrix of size $4\times 4$ consists of the composition of the transformation from $\bm{v}_1 = (\bm{o}_1, \bm{x}_1, \bm{y}_1)$ to the global reference frame $\bm{g}$ and of the transformation from the reference frame $\bm{g}$ to $\bm{v}_2 = (\bm{o}_2, \bm{x}_2, \bm{y}_2)$:
%\begin{equation}
%  \bm{T}_{v_2 v_1} =  \bm{T}_{v_2 g} \bm{T}_{g v_1}.
%  \label{eq: transformation}
%\end{equation}
%$\bm{T}_{v_2 g}$ is given by
%\begin{equation}
%  \bm{T}_{v_2 g} =
%  	\begin{bmatrix}
%            \bm{R}_{v_2 g} & \bm{o} \\
%           \bm{0}^T  & 1
%    \end{bmatrix}
%\end{equation}
%using the rotation matrix $\bm{R}_{v_2 g} = \begin{bmatrix}
%\bm{x}_2 & \bm{y}_2 & \bm{z}_2
%\end{bmatrix}$ with $\bm{z}_2 = \bm{x}_2 \times \bm{y}_2$.
%Accordingly, $\bm{T}_{g v_1}$ is computed as
%\begin{equation}
%  \bm{T}_{g v_1} = \bm{T}_{v_1 g}^{-1} = \begin{bmatrix}
%            \bm{R}^T_{v_1 g} & \bm{R}^T_{v_1 g} \left(-\bm{o}_1 \right)\\
%           \bm{0}^T  & 1
%         \end{bmatrix}.
%\end{equation}

	\subsubsection{Vehicle Bus Data}
\begin{figure*}[h]
		\centering
		\begin{subfigure}[b]{0.30\textwidth}
			\includegraphics[width=\textwidth]{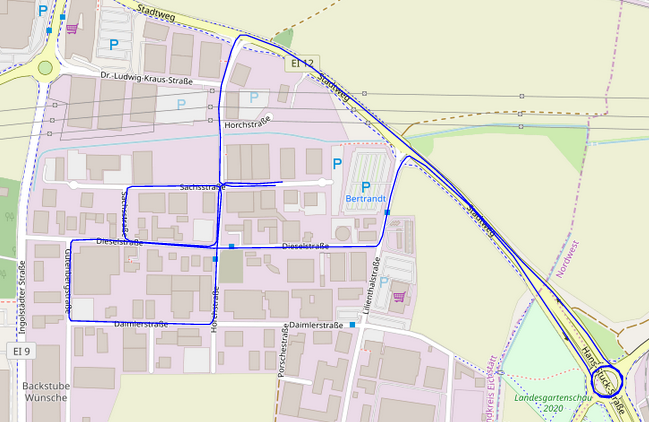}
			\caption{GPS coordinates}
		\end{subfigure}
		\begin{subfigure}[b]{0.32\textwidth}
			\includegraphics[width=\textwidth,clip=true,trim=0mm 8mm 0mm 0mm]{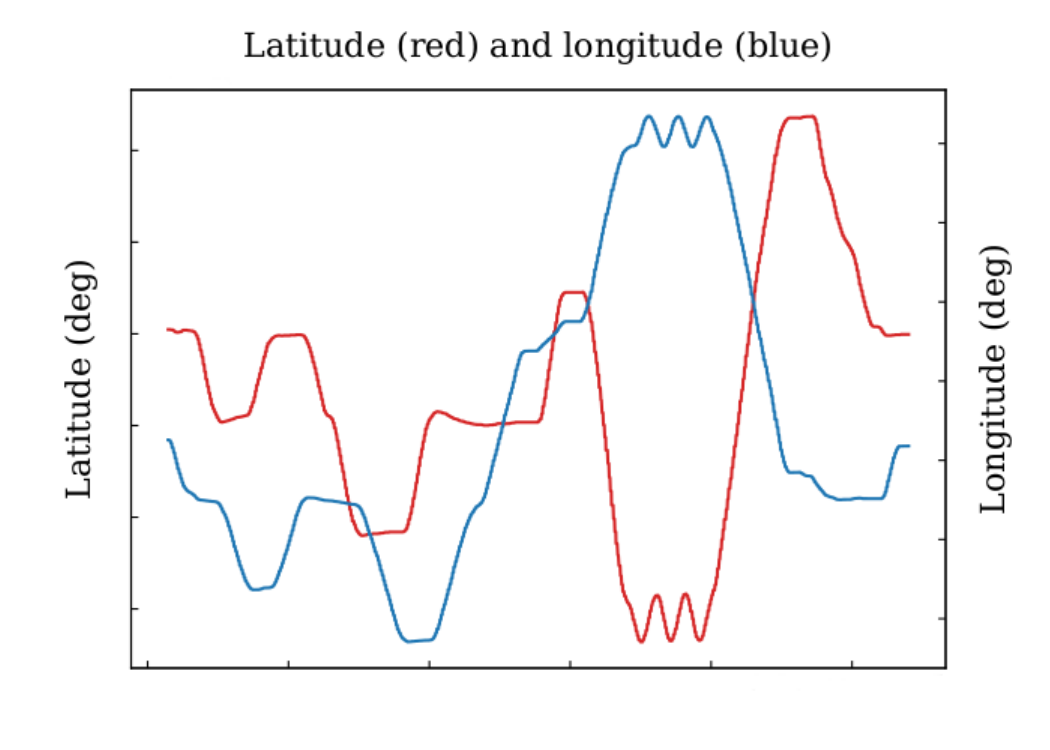}
			\caption{Degree of latitude/longitude}
		\end{subfigure}
		\begin{subfigure}[b]{0.32\textwidth}
			\includegraphics[width=\textwidth,clip=true,trim=0mm 8mm 0mm 0mm]{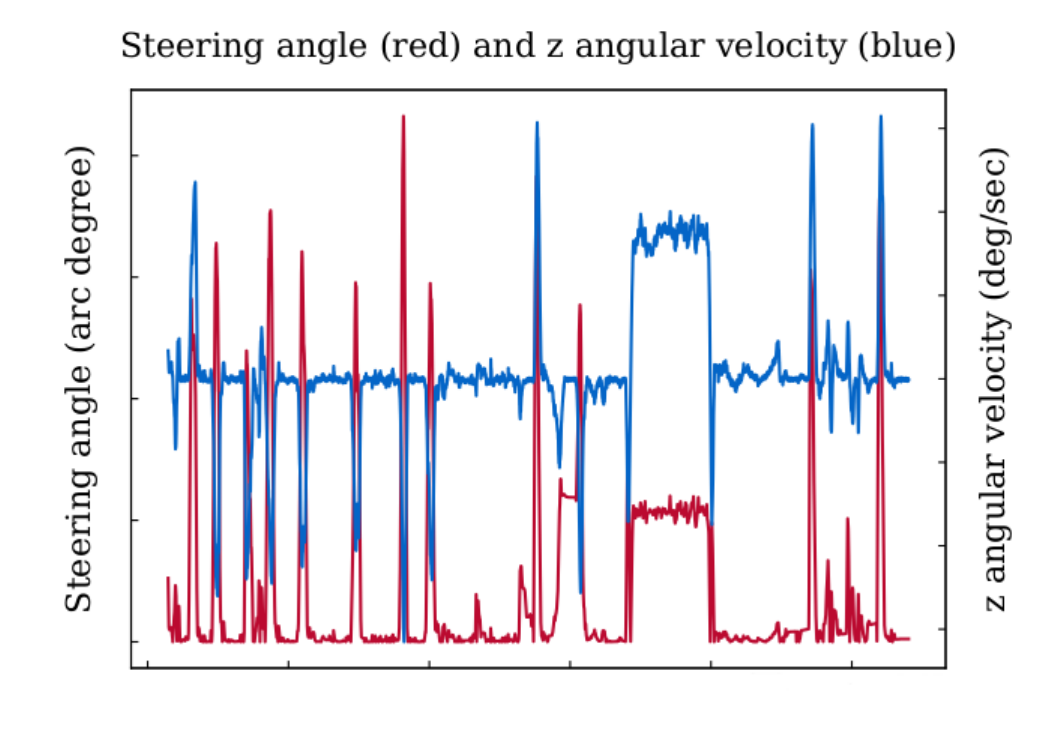}
			\caption{Steering angle and z angular velocity}
		\end{subfigure}
		\caption{Three vehicle bus signals are depicted over time. In (a), the GPS signal values are projected onto a map.
			A roundabout is visible in the lower right corner.
			This roundabout corresponds to the oscillating part of the longitude and latitude plot (b) as well as the nearly constant part in the plot of steering angle and z angular velocity (c). }
		\label{fig:busdata}
	\end{figure*}

	In addition to camera and LiDAR data, the vehicle bus data were recorded.
	It is stored in a json-file, which contains the bus signals themselves as well as the corresponding timestamps and units. 
	The signals comprise, e.g., acceleration, (angular) velocity, and GPS coordinates, brake pressure, pitch and roll angles; see Figure~\ref{fig:busdata}. 
	
    By including vehicle bus data we not only allow \aadd{} to be used for imitation (end-to-end) learning research, but also enable reinforcement learning approaches as described in \cite{Abbeel07anapplication}.
	
	\subsection{Anonymization}

    Due to recent privacy laws and regulations we utilized a state-of-the-art semantic segmentation network to blur license plates and heads of pedestrians.
    This was done for both annotated and un-annotated data (more than 400,000 images).

	\subsection{Unlabelled Sequence Data}
	\label{sec:SequenceData}
	We recorded three urban sequences in \cities{}.
	These sequences, containing closed loops, are provided as 392,556 unlabelled images (total from all six cameras), together with corresponding LiDAR and bus data.
	The sequences consist of 94,221, 164,823, and 133,512 images with corresponding timestamps, respectively.
	As mentioned in Section \ref{sec:Comparison}, these data are useful for research in end-to-end autonomous driving, depth prediction from mono/stereo images or videos \cite{Eigen2014,yang2018deep}, and SLAM.
	The latter was used for computing the LiDAR-to-camera map in Section \ref{sec:CalibrationRegistration}, see also Figure \ref{fig:MappingLiDARPoints}.
	
	\subsection{Data Labels}
	\label{sec:DataLabels}
	\subsubsection{Semantic Annotations}
    \aadd{} includes images from different road scenarios such as highway, country, and urban. 
    In total, 41,277 camera images are semantically labelled.
    Of these, 31,448 labels are for front-center camera images, 1,966 for front-left, 1,797 for front-right, 1,650 for side-left, 2,722 for side-right, and 1,694 for rear-center.
    Each pixel is assigned to a semantic class.
    
\begin{figure}
    	\centering
    	\includegraphics[width=0.46\textwidth]{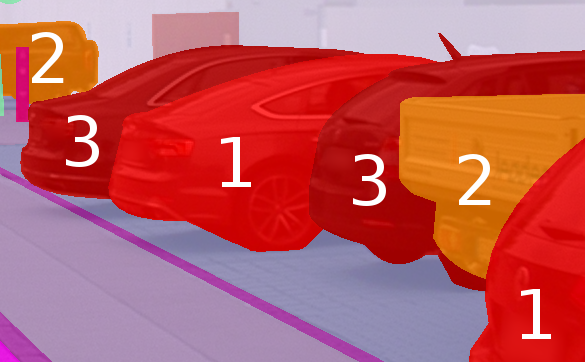}
    	\caption{Multiple adjacent car instances.  Starting from the right of the above figure, cars 1, 2, and 3 must all have different semantic classes because they share a boundary.  The next car to the left can be assigned ‘Car 1’ because it does not share a boundary with any other ‘Car 1’ instance.}
    	\label{fig:sseg_explanation}
    \end{figure}

    Where multiple instances of the same class of traffic participant (pedestrian, cyclist, car, or truck) share a boundary, they are differentiated using subclasses such as car1, car2, etc. As shown in Figure \ref{fig:sseg_explanation} this only applies to adjacent instances.
    Therefore the fact that our semantic segmentation taxonomy has classes car[1-4] does not imply that there is a maximum of 4 cars per image.
    
	In total, there are 38 classes in our semantic segmentation schema.
	The lower panel of Figure \ref{fig:class_instances} shows these classes along with the distribution of pixels in our dataset.
	The number of instances of traffic participants annotated with semantic labels in our dataset are depicted in the upper-right panel of Figure \ref{fig:class_instances}.
	Predictably, the traffic participant instance counts are dominated by cars, trucks, and pedestrians.
%	\textbf{LiDAR data for how many frames?}

    We also provide LiDAR point clouds for 38,481 semantically labelled images.
    The 3D semantic labels are obtained by mapping the 3D points to the semantic segmentation images using the LiDAR to camera mapping described in Section \ref{sec:CalibrationRegistration}.
    
    \subsubsection{Instance Segmentation Annotations}
    We generated instance segmentation annotations from all semantic segmentation annotations. These instance annotations are available for all classes which represent traffic participants such as pedestrians, cars, etc.
    %Due to occlusions etc. this conversion can sometimes be non-trivial.
    %To deal with this, ad hoc solutions were implemented based on the edge cases encountered during visual inspection.
    
	\subsubsection{3D Bounding Boxes}

	\begin{figure*}[htp]
	\centering
	\begin{tabular} {cc}
		\includegraphics[width=0.36\linewidth,clip=true,trim = 0mm 0mm 10mm 5mm]{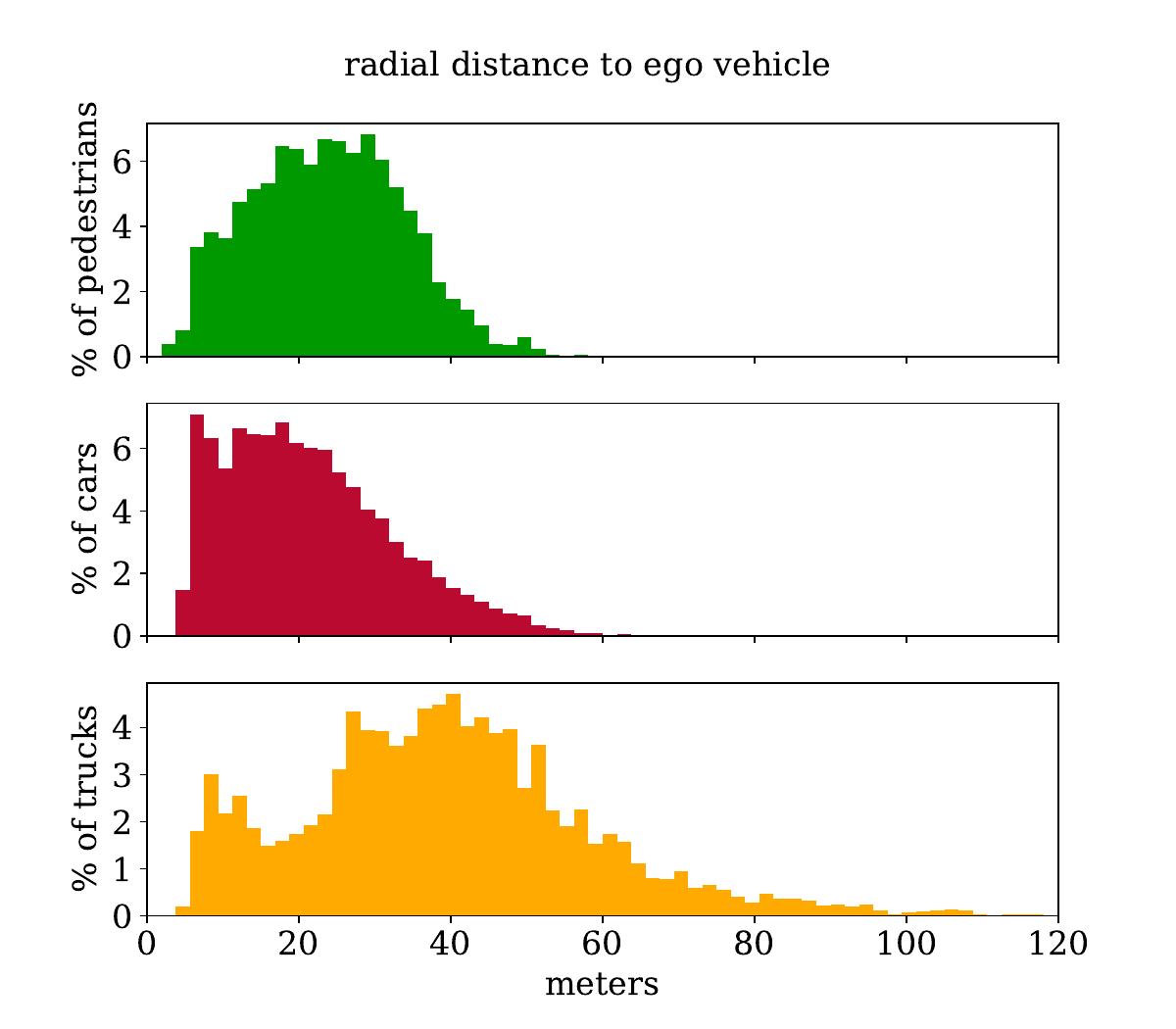} & \includegraphics[width=0.36\linewidth,clip=true,trim = 0mm 0mm 10mm 5mm]{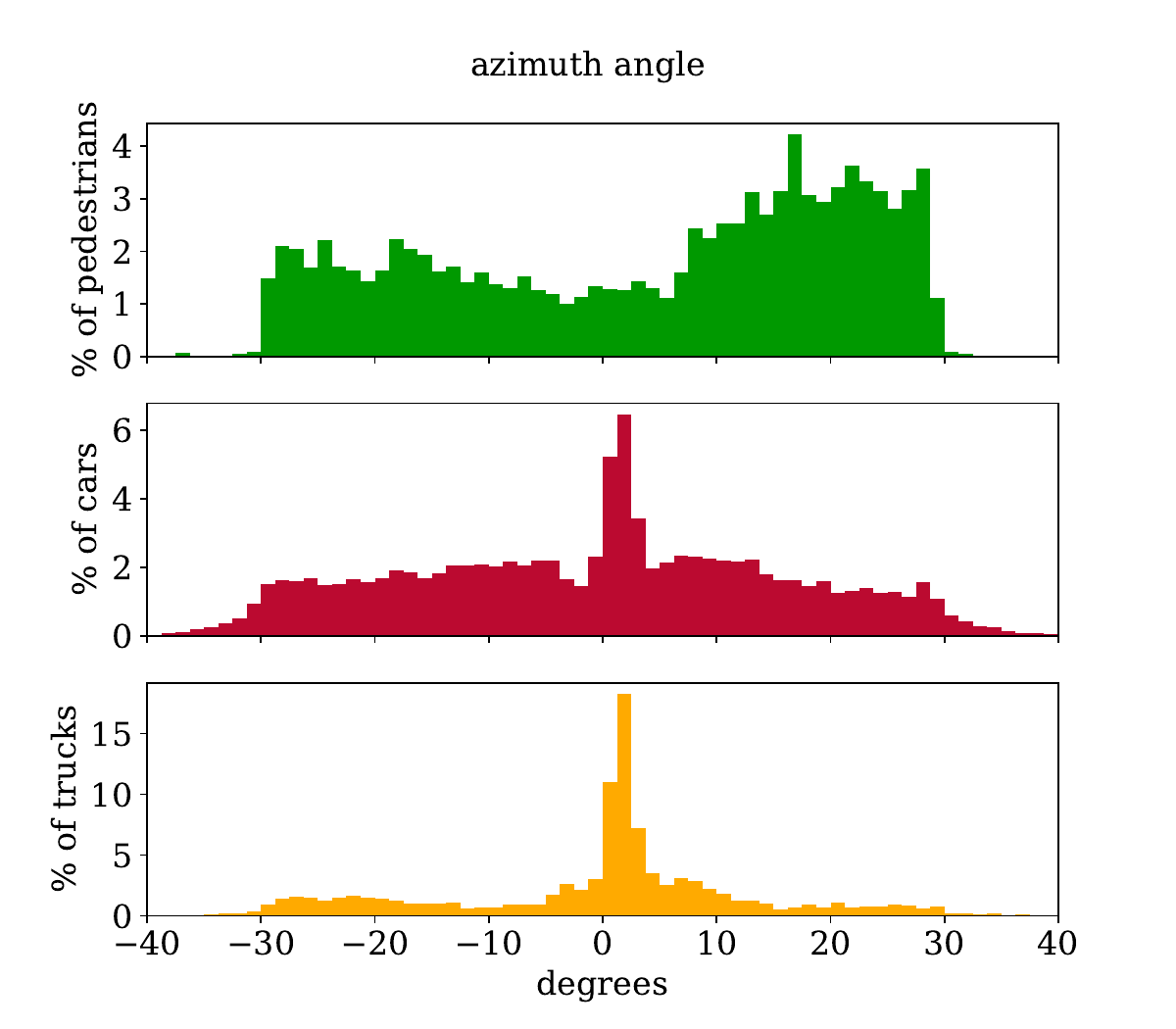}  %\\
%		(a) & (b)
	\end{tabular}
\vspace*{-2ex}
	\caption{Distributions of radial distances and azimuthal angles for pedestrian, car, and truck bounding box object classes}
	\label{fig:radial_azimuthal_distributions}
	\end{figure*}
	\begin{figure*}[htp]
		\centering
		\includegraphics[width=0.34\linewidth]{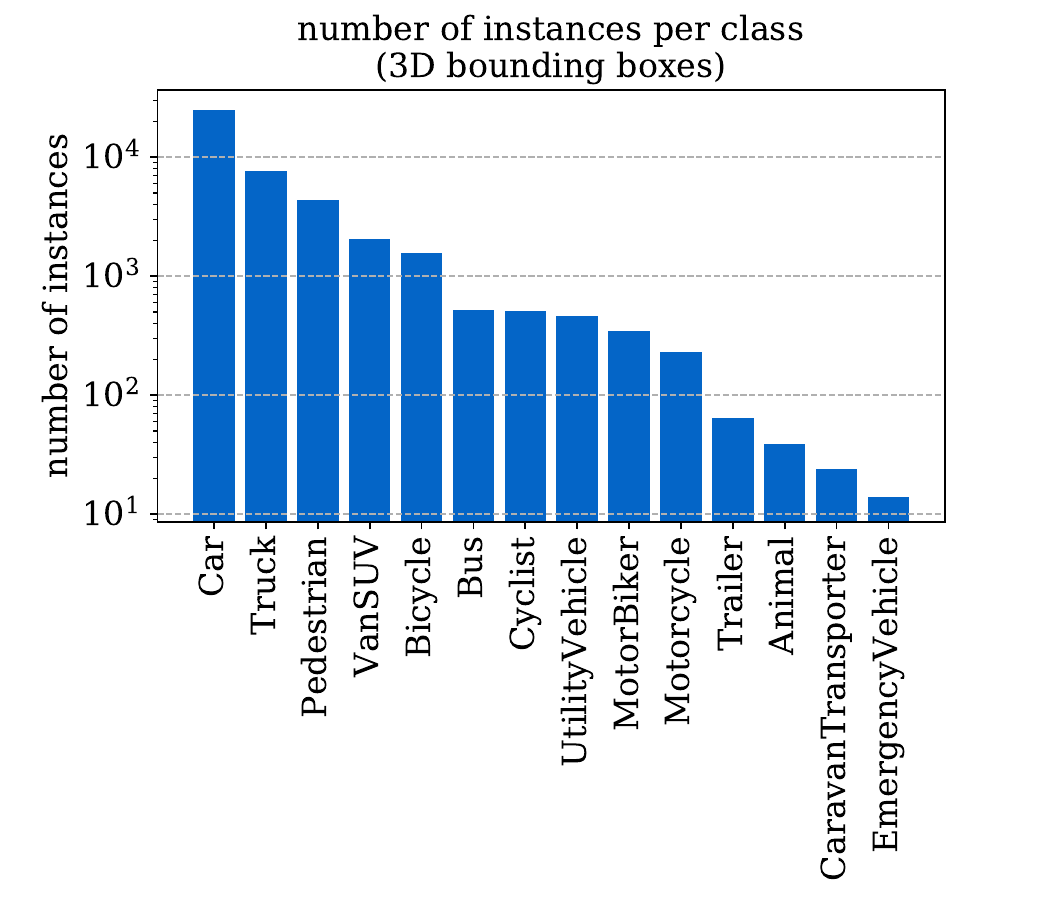}
		\includegraphics[width=0.34\linewidth]{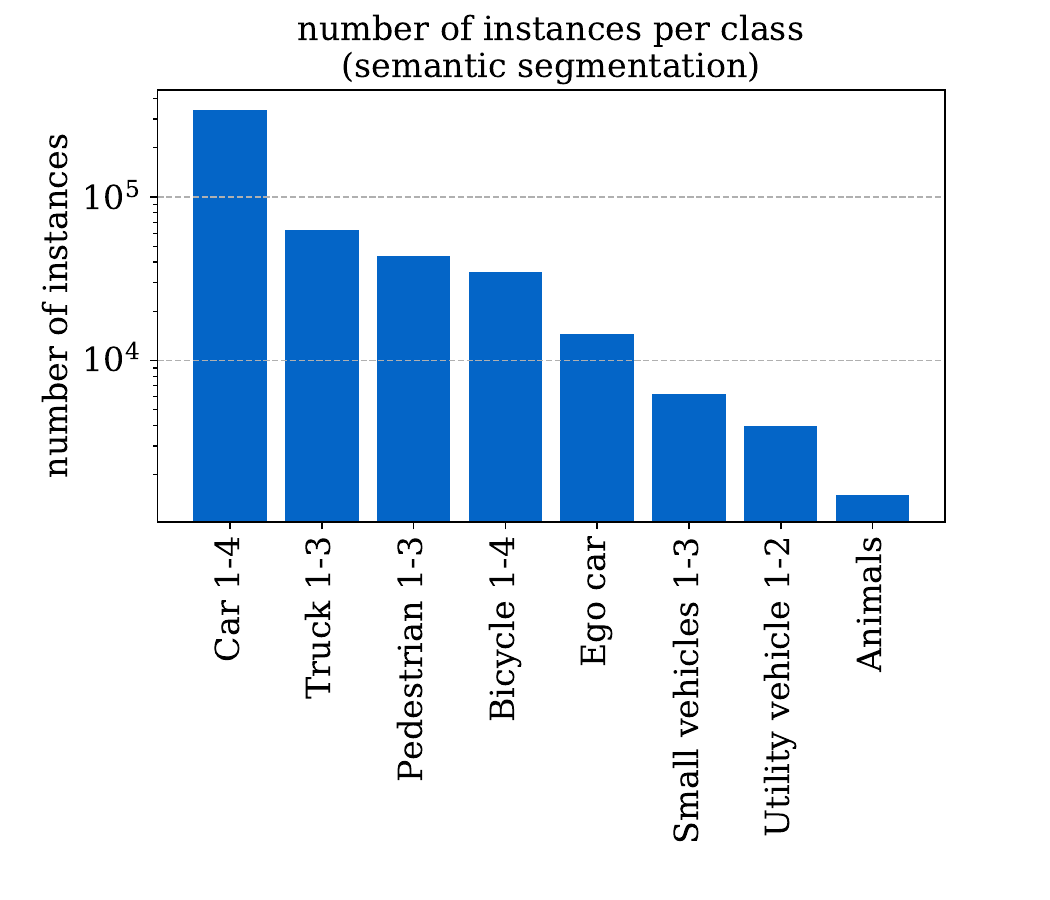} \\[-1.0ex]
		\includegraphics[width=0.70\linewidth]{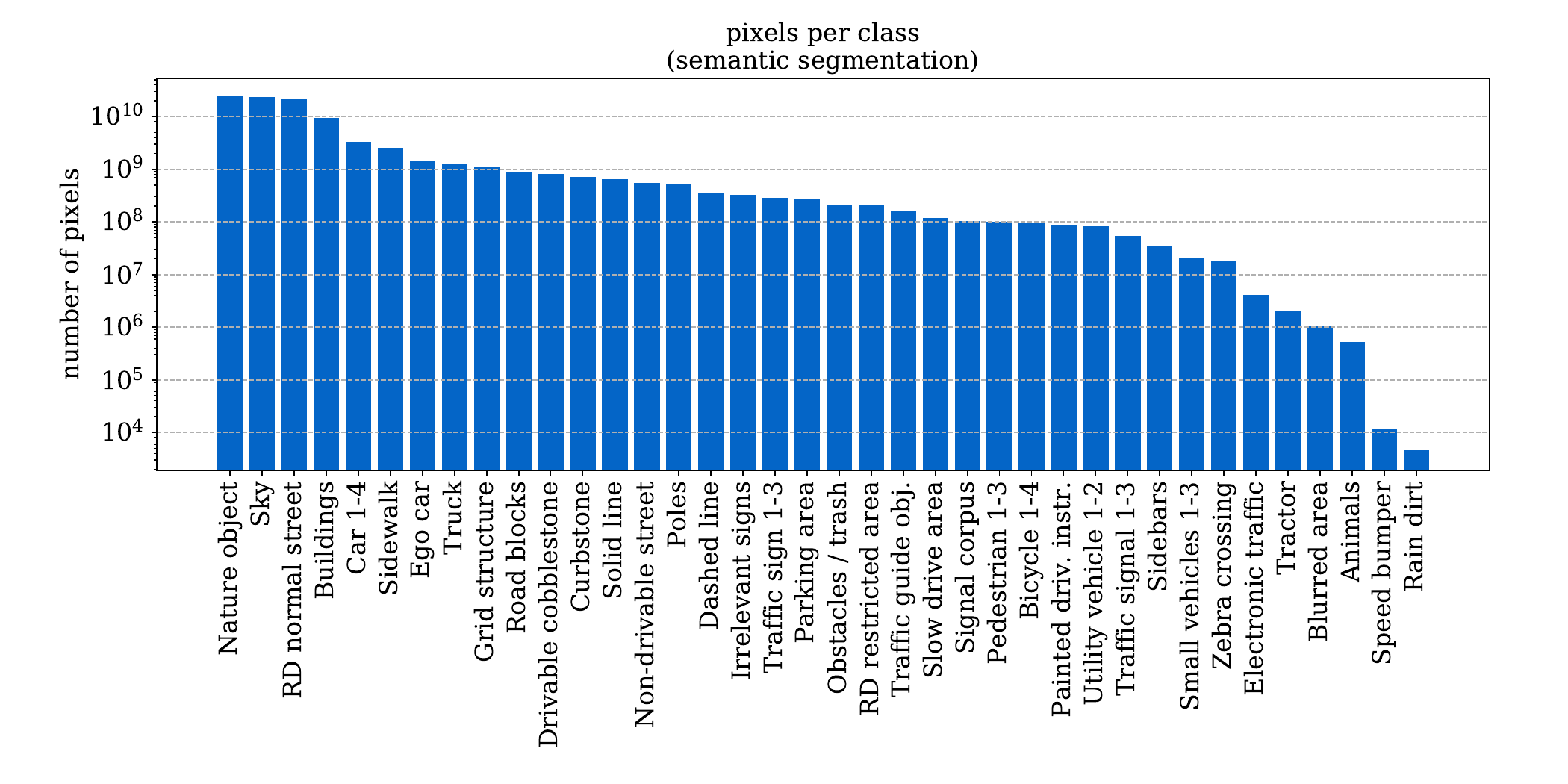} \\[-2ex]
		\caption{Number of 3D bounding box instances per class, number of traffic participant instances per class (semantic segmentation), semantic segmentation pixel counts per class. Note that ``Cyclist'' and ``MotorBiker'' stand for a person including the corresponding vehicle.}
		\label{fig:class_instances}
	\end{figure*}
	For 12,497 of the front-camera frames with semantic annotation, we also provide 3D bounding boxes for a variety of vehicles, pedestrians, and other relevant objects.  
 	Figure \ref{fig:class_instances} (top left) shows the full list of annotated classes, along with the number of instances of each class in \aadd{}.
	Where objects are partially occluded, human annotators estimated the bounding box to the best of their ability.
	Examples of this can be seen in the second panel of Figure \ref{fig:headline}.

	3D bounding boxes were annotated in LiDAR space.
	To do so, we first combined the point clouds from all LiDARs, then culled them to the view frustrum of the front-center camera.
	Therefore, we provide 3D bounding boxes for the points within the field of view of the front-center camera.
	%Nevertheless, individual points used for annotation could originate from any LiDAR sensor, not only the front-center LiDAR.

	LiDAR point clouds are sparse relative to images. As a result, distant or small objects may not be represented by (m)any points.
	Since the 3D bounding boxes are derived from LiDAR point clouds, %this sparsity means that 
	objects may be visible in images but lack corresponding 3D bounding boxes.

	Figure \ref{fig:radial_azimuthal_distributions} (left) shows the distributions of the radial distances to the ego vehicle of the 3D bounding boxes in our dataset for three classes: pedestrians, cars, and trucks.
	As one may expect, trucks, being physically larger than cars and pedestrians, have a higher optical cross section, and are thus seen and annotated in the LiDAR point clouds to a farther distance.
	
	Although cars are metallic and larger than pedestrians, Figure \ref{fig:radial_azimuthal_distributions} (left) shows that the distribution of annotated cars drops off quicker with distance than the distribution of annotated pedestrians.
	A possible explanation for this would be that the distributions of the annotations largely reflect the real world distributions, i.e., the percentage of cars at long range really is smaller than the percentage of pedestrians at long range. 
	This seems unlikely since \aadd{} contains recordings from highways and country roads, where we expect many distant cars.
	A more plausible explanation relies on the fact that LiDAR visibility is not determined solely by optical cross section, but also by occlusion, which is affected by the size, shape, and spatial distribution of objects. 
%	Since a car is larger than a pedestrian, the solid angle it occludes is larger than the solid angle that a pedestrian at the same distance would occlude.
	Since cars are generally on roads, they tend to be almost directly ahead of the ego vehicle (see right panels of Figure \ref{fig:radial_azimuthal_distributions}).
	Thus they often occlude other cars rather than pedestrians, 
%	not only do cars occlude a larger solid angle, but the occluded region is more likely to contain other cars than pedestrians, since
 the azimuthal distribution of which is more even as shown in Figure \ref{fig:radial_azimuthal_distributions} (right).
	This might explain why the distribution of annotated cars falls off quicker with distance than the one for pedestrians. 
	
	The same argument does not hold for trucks, despite the fact that they are also mostly directly ahead of the ego vehicle (see Figure \ref{fig:radial_azimuthal_distributions}).
	They are far less common than cars in \aadd{} as depicted in the upper panels of Figure \ref{fig:class_instances} (note the logarithmic $y$-axes). 
	This suggests that potential occluding objects are usually not other trucks.
	That being the case, since trucks are wider and taller than other vehicles, they are less likely to be fully occluded so that they can be perceived to a farther distance. 

	\subsection{Tutorial}
	Since our aim is to provide a valuable resource to the community, it is important that \aadd{} be easy to use.
	To this end we provide a Jupyter Notebook tutorial with the download, which details how to access and use the dataset. 

	%all points within one lidar rotation

%	\begin{tiny}
%	\begin{table} [htp]
%		\centering
%		\begin{footnotesize}
%			\begin{tabular}{rll}
%				%\hline
%				Class & Instance count\\
%				\hline %\hline
%				Animal			&	00000\\
%				Car				&	11111\\
%				Pedestrian		&	22222\\
%				Tractor 		&	33333\\
%				Truck 			&	44444\\
%				Utility Vehicle &	55555\\
%				Cyclist			&	66666\\
%				Bicycle			&	77777\\
%				Bus				&	88888\\
%				Motorbiker		&	99999\\
%				Motorcycle		&	10101\\
%				Trailers		&	11111\\
%				Van/SUV			&	12121\\
%				Emergency vehicle &	999\\
%				\hline
%			\end{tabular}
%		\end{footnotesize}
%		\caption{\textbf{---ac:TODO: fill in real numbers or replace with graph?---}Class list and instance counts for 3D bounding boxes}
%		\label{tbl:3d_bb_class_list}
%	\end{table}
%	\end{tiny}

	\section {Experiment: Semantic Segmentation}
	\label{sec:experiments}

	\begin{figure*}[ht]
		\centering
		\begin{tabular} {ccc}
			\includegraphics[width=0.26\linewidth]{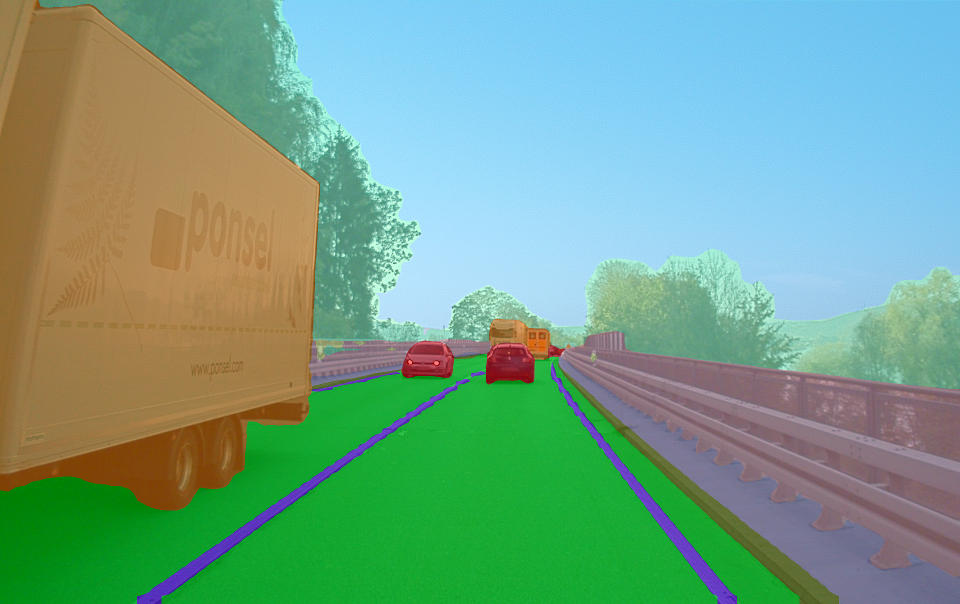} & \hspace*{-1.1em} \includegraphics[width=0.26\linewidth]{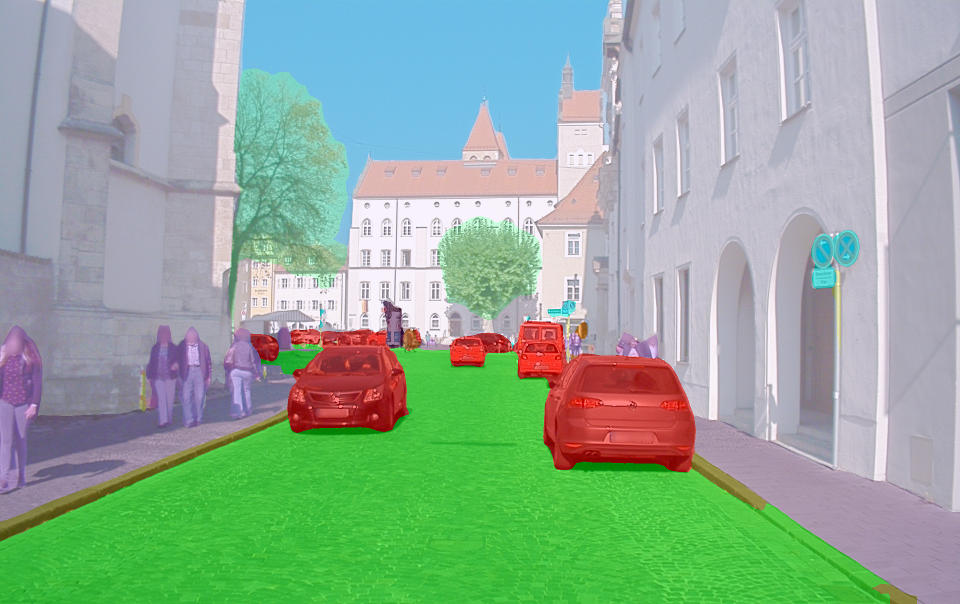} & \hspace*{-1.1em}
			\includegraphics[width=0.26\linewidth]{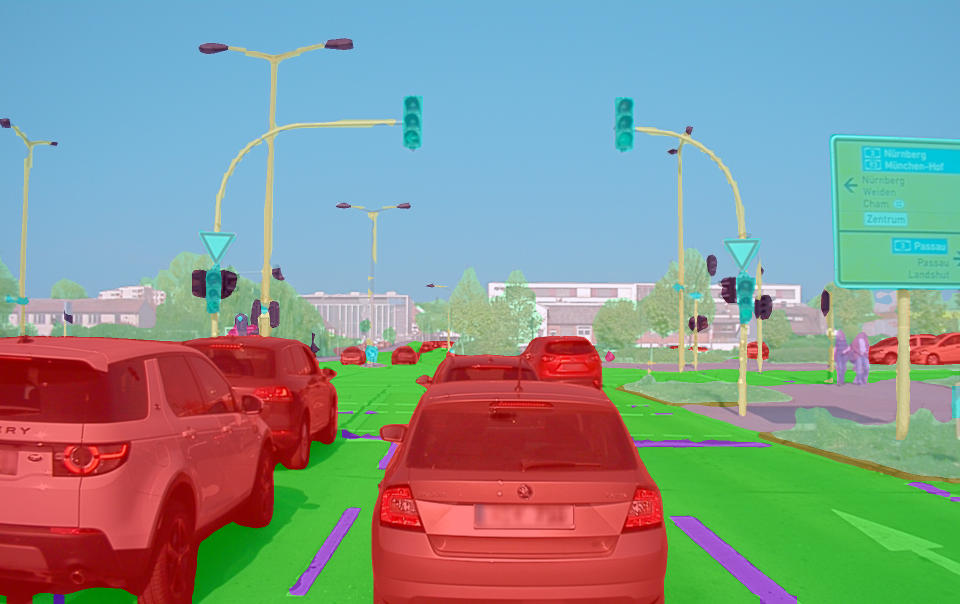} \\[0.0ex]
			\includegraphics[width=0.26\linewidth]{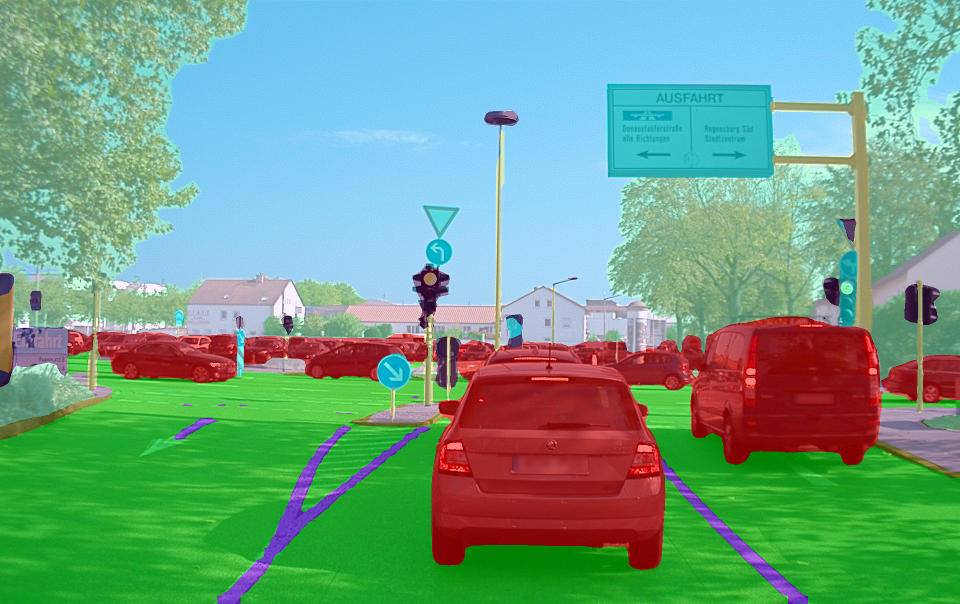} & \hspace*{-1.1em}
			\includegraphics[width=0.26\linewidth]{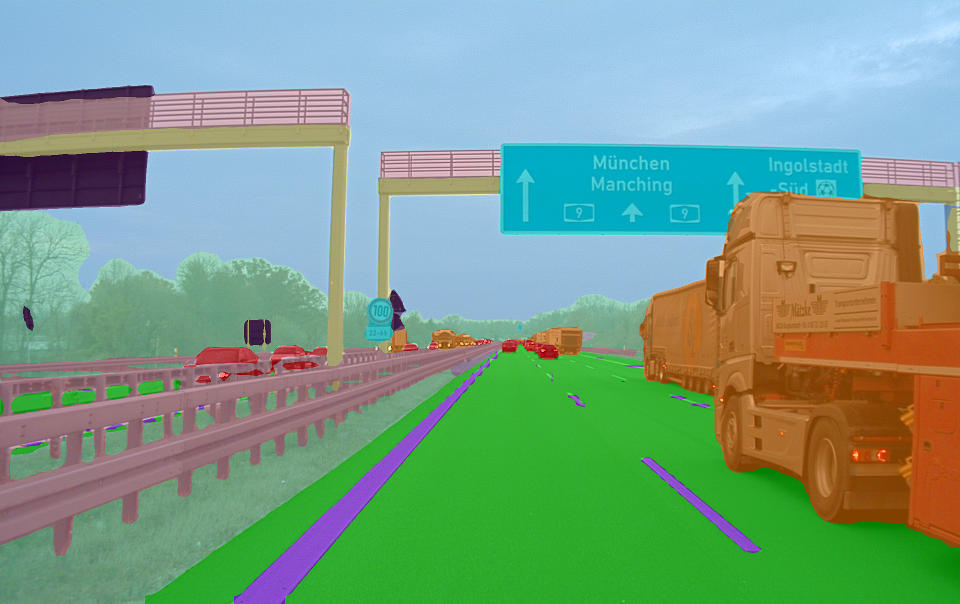} & \hspace*{-1.1em} \includegraphics[width=0.26\linewidth]{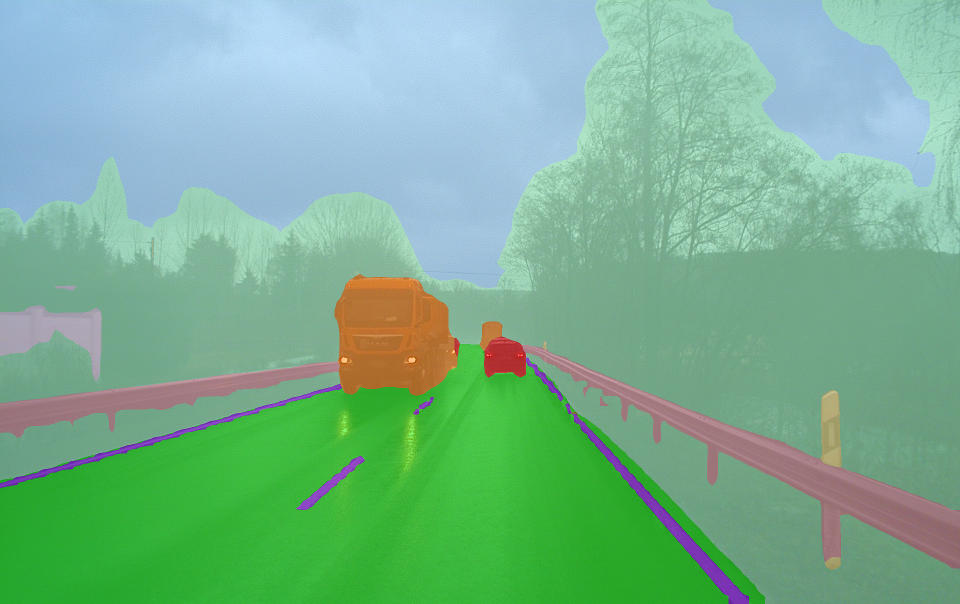}
		\end{tabular}
		\caption{Visual result on test set images of different scenarios}
		\label{fig:segmentation_visual_results}
	\end{figure*}

	In this section we present the results of training and evaluating a semantic segmentation network on \aadd{}. 
	In order to establish baseline results, our experiment follows state-of-the-art methods \cite{long2014fully,deeplabv3plus2018}, training a fully convolutional network to classify each pixel within an image.
	
	For our experiments, we used 40,030 RGB images with a resolution of $1920\times 1208$ pixels.
	We split the data into train (28,015 images), validation (4,118), and test (7,897) sets.
	The experiments were conducted for 18 classes of interest, and an additional background class.
	These 19 classes were chosen to be as similar as possible to the Cityscapes taxonomy, and were generated by merging similar classes in our 38-class taxonomy.
	Random cropping, brightness, contrast, and flipping augmentations were applied.
	As is standard practice, we use mean Intersection over Union (IoU) to evaluate the performance of the model.

    \subsection{Baseline Results} \label{subsec:baseline_results}
    \renewcommand{\arraystretch}{1.5}
%    \begin{table} 
%    	\centering
%    	\footnotesize
%    	\begin{tabular}{ lc  }
%    		Hyperparameter & Value\\
%    		\hline		
%    		Optimizer & SGD + momentum \\
%    		Learning rate & $1.0 \times 10^{-2}$  \\
%    		Learning rate schedule & Polynomial decay\\
%    		Momentum & 0.9 \\
%    		\hline  
%    	\end{tabular}
%    	\caption{Semantic segmentation network hyperparameters}
%    	\label{tbl:segmentation_hyperparams}
%    \end{table}

	The network architecture used ResNet-101 \cite{he2016identity} as the encoder and the pyramid scene parsing network \cite{zhao2017pspnet} module as the decoder.
	%Table \ref{tbl:segmentation_hyperparams} shows the network hyperparameters used for our experiments.
	The encoder %of the network 
	was initialized with weights from ImageNet pre-training.
	We trained the network using stochastic gradient descent with momentum.
	The initial learning rate was 0.01 and the momentum parameter was 0.9.
	The learning rate was decreased polynomially during training. 
	The feature map of the last encoder layer has spatial dimensions $\frac{1}{16}$ those of the input image.
	This model achieves a mean IoU over the 18 foreground classes of 71.01\% on the test set, as shown in Table \ref{tbl:segmentation_results}.
	Figure \ref{fig:segmentation_visual_results} shows some visual examples of the network output.
	
		\renewcommand{\arraystretch}{1.5}
		\begin{table} 
			\centering
			\footnotesize
			\begin{tabular}{ lc  }
				Architecture/Training & Mean IOU  \\
				\hline		
				Baseline (ResNet-101 + PSP-Net) & 71.01\% \\ 
				With pre-trained weights (ResNet-50 + PSP-Net) & 68.40\% \\ 
				Without pre-trained weights (ResNet-50 + PSP-Net) & 65.31\% \\ 
				With anonymized images (ResNet-101 + PSP-Net) & 70.94\% \\
				\hline  
			\end{tabular}
			\caption{Evaluation results on the test set} \label{tbl:segmentation_results}\vspace*{-2ex}
		\end{table}

\subsection{Usage of Pre-Trained weights} \label{subsec:with_without_pretrained_weights}
    We assess the influence of ImageNet pre-training versus random weight initialization.
    The network architecture is similar to the baseline experiment, but the encoder module is replaced by ResNet-50, and the final encoder layer spatial dimensions are $\frac{1}{8}$ those of the input image.
    The training algorithm and hyperparameters are the same as for the baseline experiments with the only exception that the initial learning rate is $1.0 \times 10^{-4}$ for the network with pre-trained weights.
    The mean IoU evaluation results are shown in Table \ref{tbl:segmentation_results}.
    The model with pre-trained weights achieves a better result.

\subsection{Training With Anonymized Images}
    The results which we have discussed thus far apply to models trained on images which were not anonymized.
    Since legal requirements require that our dataset be anonymized prior to public release, we investigate the effect of anonymization on performance.
    To do this we trained our network with anonymized images, where faces and vehicle number plates were blurred.
    The network architecture and experimental setup are the same as in the baseline experiment, with the encoder once again initialized with ImageNet pre-trained weights.
    The model achieves a mean IoU of 70.94\% on the test set, which is very similar to the baseline result, and does not immediately suggest that anonymization has an adverse effect on the semantic segmentation task.
    Table \ref{tbl:segmentation_results} shows the results of all of our experiments.

	\section {Conclusions and Outlook}
	We provide a commercially usable dataset, which includes camera, LiDAR and bus data recorded from a \car{}. 
	The data from six cameras and five LiDAR sensors are registered to a global reference frame and include precise timestamps. 
	Rich data is provided, in particular full $360\degree$ sensor coverage of the vehicle environment.
	We have strived to make \aadd{} as accessible and easy to use as possible (license, privacy concerns, interactive tutorial), with the end goal of advancing state-of-the-art commercial and academic research in computer vision, machine learning, and autonomous driving.
	
	We expect to continuously update \aadd{} in line with current frontiers in research.
	Indeed, instance segmentation annotations were not included in the initial public release, but are now available for download.  
	Furthermore we plan to define benchmarks and challenges to allow researchers to easily and fairly compare their algorithms.
	To this end, we have labeled a test set of $\sim$10K images with semantic segmentation annotations, and are currently exploring how best to allow the community to benchmark against this ground truth.
	% ems and provide a tes,t framework depending on the current needs of the research community. 

\ifcvprfinal

   \section {Acknowledgments}

   We would like to thank Josua Schuberth, Sirinivas Reddy Mudem, Kevin Michael
   Bondzio, Yunlei Tang, Ajinkya Khoche, Christopher Schmidt, Sumanth Venugopal
   and Sah Surendra for their help in checking the quality of the dataset. In 
   addition, we would like to thank E.S.R. Labs AG for their help in
   developing this dataset.

\else{}\fi

%\nocite{*}
\bibliographystyle{unsrt}
\bibliography{ms}
\end{document}